\documentclass[letterpaper]{article}
\usepackage{aaai2026}
\nocopyright
\usepackage{times}
\usepackage{helvet}
\usepackage{courier}
\usepackage[hyphens]{url}
\usepackage{graphicx}
\usepackage[numbers]{natbib}
\usepackage{caption}
\usepackage{amsmath}
\usepackage{amsfonts}
\usepackage{algorithm}
\usepackage{algorithmic}
\usepackage{booktabs}
\usepackage{multirow}

\title{AgentSHAP: Interpreting LLM Agent Tool Importance\\ with Monte Carlo Shapley Value Estimation}

\author{
    Miriam Horovicz\\
    Fiverr Labs\\
    miriam@f-labs.io
}

\begin{document}
\maketitle

\begin{abstract}
LLM agents that use external tools can solve complex tasks, but understanding which tools actually contributed to a response remains a blind spot. No existing XAI methods address tool-level explanations. We introduce \textbf{AgentSHAP}, the first framework for explaining tool importance in LLM agents. AgentSHAP is model-agnostic: it treats the agent as a black box and works with any LLM (GPT, Claude, Llama, etc.) without needing access to internal weights or gradients. Using Monte Carlo Shapley values, AgentSHAP tests how an agent responds with different tool subsets and computes fair importance scores based on game theory. Our contributions are: (1) the first explainability method for agent tool attribution, grounded in Shapley values from game theory; (2) Monte Carlo sampling that reduces cost from $O(2^n)$ to practical levels; and (3) comprehensive experiments on API-Bank showing that AgentSHAP produces consistent scores across runs, correctly identifies which tools matter, and distinguishes relevant from irrelevant tools. AgentSHAP joins TokenSHAP (for tokens) and PixelSHAP (for image regions) to complete a family of Shapley-based XAI tools for modern generative AI.

Code:~\mbox{\url{https://github.com/GenAISHAP/TokenSHAP}}.
\end{abstract}

\section{Introduction}

Large Language Model (LLM) agents can now use external tools calculators, databases, web search, and APIs \cite{schick2023toolformer, qin2023tool}. This makes them far more capable than standalone LLMs. An agent with access to a calculator can solve math problems accurately. An agent with stock market APIs can provide real-time financial information. An agent with Wikipedia access can answer factual questions with citations.

But with this power comes a new challenge: \textit{Which tools actually matter for a given response?}

This question matters for two key reasons. First, \textbf{optimization}: system designers need to decide which tools to include, since more tools mean higher costs and latency. Logs show which tools were called, but not how much each tool actually mattered for the response. A tool might be called frequently yet contribute little. Second, \textbf{understanding}: developers need to verify that the right tools are contributing. If a Calculator tool has low importance on a math query, something is wrong with the agent's behavior.

Current XAI methods don't help here. They focus on input tokens \cite{goldshmidt2024tokenshap}, attention patterns \cite{vig2019analyzing}, or feature importance for tabular data \cite{lundberg2017unified}. To our knowledge, no existing explainability framework addresses tool-level attribution in LLM agents.

We introduce \textbf{AgentSHAP}, the first framework that explains tool importance in LLM agents. A key advantage is that AgentSHAP is model-agnostic: it treats the agent as a black box, requiring only input-output access. This means it works equally well with local models you run yourself or third-party APIs, without needing internal weights, gradients, or architecture details.

\begin{figure}[t]
    \centering
    \includegraphics[width=\columnwidth]{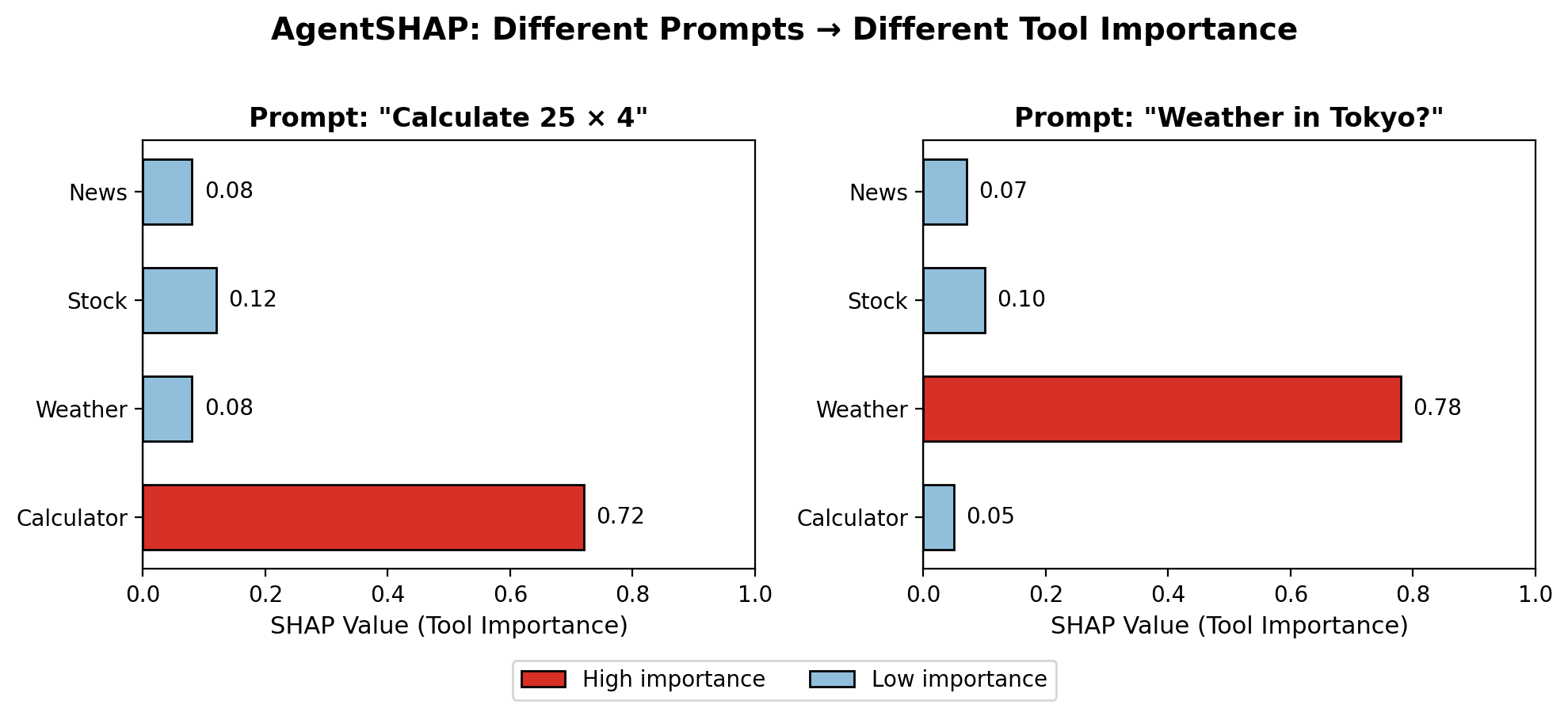}
    \caption{\textbf{AgentSHAP shows which tools matter.} Different prompts activate different tools. For a math query, Calculator scores highest (0.72). For a weather query, Weather tool scores highest (0.78). Red bars indicate high importance, blue bars indicate low importance.}
    \label{fig:teaser}
\end{figure}

The key idea is simple: run the agent with different subsets of tools and see how the response changes. Tools that cause big changes when removed are important; tools that don't matter get low scores. Figure~\ref{fig:teaser} shows example results.

We make the following contributions:
\begin{enumerate}
    \item \textbf{AgentSHAP framework}: We propose the first method to compute tool importance scores in LLM agents using Shapley values. The scores satisfy key game-theoretic properties: efficiency, symmetry, and null player.

    \item \textbf{Efficient Monte Carlo estimation}: Computing exact Shapley values requires $2^n$ evaluations. We use Monte Carlo sampling to reduce this to practical levels while maintaining accuracy.

    \item \textbf{Comprehensive evaluation}: We test AgentSHAP on the API-Bank benchmark with real executable tools across four experiments: consistency, faithfulness, irrelevant tool injection, and cross-domain attribution.

    \item \textbf{Open-source release}: AgentSHAP is available as part of the TokenSHAP library at \url{https://github.com/GenAISHAP/TokenSHAP}.
\end{enumerate}

\section{Related Work}

\textbf{LLM Agents with Tools.} The capability of LLMs to use external tools has been a major research focus \cite{mialon2023augmented, wang2023survey, xi2023rise}. Toolformer \cite{schick2023toolformer} pioneered self-supervised tool learning, while ReAct \cite{yao2022react} introduced interleaved reasoning and acting. Gorilla \cite{patil2023gorilla} demonstrated accurate API calling, and HuggingGPT \cite{shen2023hugginggpt} orchestrates multiple AI models as tools. Recent work on autonomous agents \cite{shinn2023reflexion} shows agents can learn from feedback. Benchmarks like API-Bank \cite{li2023apibank} and ToolBench \cite{qin2023toolllm} provide standardized evaluation with real executable APIs. Despite this progress, no existing work addresses explaining \textit{which} tools contributed to agent responses.

\textbf{Shapley Values for Explainability.} Shapley values from cooperative game theory \cite{shapley1953value} provide a principled, axiomatic approach to fair attribution. SHAP \cite{lundberg2017unified} popularized this for machine learning feature importance, with extensions to tree models \cite{lundberg2020local}. The key challenge is computational: exact Shapley values require $O(2^n)$ coalition evaluations. Monte Carlo sampling \cite{castro2009polynomial} addresses this by estimating values through random permutations, trading exactness for efficiency.

\textbf{Explainability for LLMs.} Explaining LLM behavior remains challenging. Attention visualization \cite{vig2019analyzing} shows what tokens the model attends to, but attention doesn't always reflect importance. Integrated Gradients \cite{sundararajan2017axiomatic} and LIME \cite{ribeiro2016lime} provide input attribution but require model access. TokenSHAP \cite{goldshmidt2024tokenshap} uses Monte Carlo Shapley values to explain which input tokens matter for LLM outputs in a model-agnostic way. PixelSHAP \cite{goldshmidt2025pixelshap} extends this to vision-language models. AgentSHAP completes this family by explaining tool importance a new dimension of explainability unique to agentic systems.

\section{Method}

\begin{figure*}[t]
    \centering
    \includegraphics[width=0.95\textwidth]{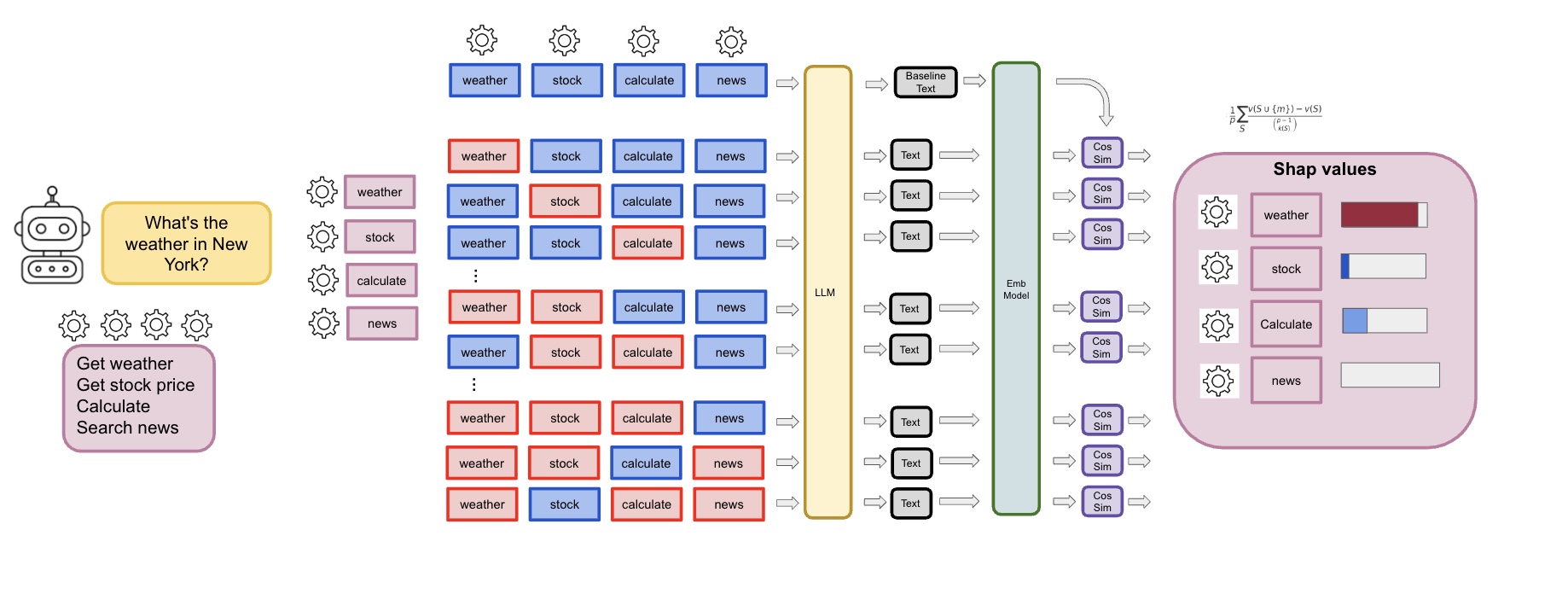}
    \caption{\textbf{How AgentSHAP works.} (1) Run the agent with all tools to get a baseline response. (2) Test the agent with different tool subsets using Monte Carlo sampling. (3) Compute Shapley values measuring each tool's contribution to response quality. The value function uses semantic similarity (cosine similarity on embeddings) to compare responses.}
    \label{fig:method}
\end{figure*}

An LLM agent $\mathcal{A}$ has access to a set of tools $T = \{t_1, t_2, \ldots, t_n\}$. Each tool $t_i$ has a name, description, and executable function. Given a user prompt $p$, the agent produces a response $r = \mathcal{A}(p, T)$ by potentially calling one or more tools. Figure~\ref{fig:method} illustrates the overall approach.

Our goal is to compute importance scores $\phi = \{\phi_1, \phi_2, \ldots, \phi_n\}$ where $\phi_i$ quantifies how much tool $t_i$ contributed to the final response. These scores should be fair (tools that contribute equally get equal scores), complete (accounting for all contribution), and interpretable (higher scores mean more importance).

\subsection{Shapley Values for Tools}

We formulate tool attribution as a cooperative game where the tools $T = \{t_1, \ldots, t_n\}$ are the players. A coalition is any subset of tools $S \subseteq T$, and the value function $v(S) = \text{sim}(\mathcal{A}(p, S), \mathcal{A}(p, T))$ measures how similar the response with only tools $S$ is to the full response with all tools $T$. We use cosine similarity on text embeddings (text-embedding-3-large) to capture semantic meaning rather than surface-level word overlap.

The Shapley value for tool $t_i$ is:
\begin{equation}
\phi_i = \sum_{S \subseteq T \setminus \{t_i\}} \frac{|S|!(n-|S|-1)!}{n!} [v(S \cup \{t_i\}) - v(S)]
\label{eq:shapley}
\end{equation}

This formula averages the marginal contribution of tool $t_i$ across all possible orderings in which tools could be added. The weighting ensures fairness: each ordering is equally likely. Shapley values satisfy key axioms (efficiency, symmetry, null player, linearity) that make them the unique fair attribution method \cite{lundberg2017unified}.

\subsection{Monte Carlo Estimation}

Computing exact Shapley values requires evaluating $2^n$ coalitions, which is infeasible for agents with many tools. We use Monte Carlo sampling to estimate values efficiently.

\begin{algorithm}[t]
\caption{Monte Carlo Shapley for Tools}
\label{alg:mc_shapley}
\begin{algorithmic}[1]
\REQUIRE Tools $T$, prompt $p$, sampling ratio $\rho$
\ENSURE Shapley values $\phi_1, \ldots, \phi_n$
\STATE Get baseline response: $r_{\text{base}} \leftarrow \mathcal{A}(p, T)$
\STATE Initialize: $\phi_i \leftarrow 0$ for all $i$
\FOR{each tool $t_i \in T$}
    \STATE Compute leave-one-out: $v(T \setminus \{t_i\})$
\ENDFOR
\STATE $m \leftarrow \lfloor \rho \cdot (2^n - n - 1) \rfloor$
\FOR{$j = 1$ to $m$}
    \STATE Sample random subset $S \subset T$
    \STATE Get response: $r_S \leftarrow \mathcal{A}(p, S)$
    \STATE Compute similarity: $v(S) \leftarrow \text{sim}(r_S, r_{\text{base}})$
    \STATE Update marginal contributions for each tool
\ENDFOR
\STATE Compute final $\phi_i$ from Equation~\ref{eq:shapley}
\RETURN $\phi_1, \ldots, \phi_n$
\end{algorithmic}
\end{algorithm}

Our algorithm (Algorithm~\ref{alg:mc_shapley}) has two phases:
\begin{enumerate}
    \item \textbf{Leave-one-out}: We always test removing each tool individually. This ensures every tool's direct effect is measured.
    \item \textbf{Random sampling}: We sample additional random coalitions to capture tool interactions and improve estimates.
\end{enumerate}

With sampling ratio $\rho$, we evaluate approximately $n + \rho \cdot (2^n - n - 1)$ coalitions instead of $2^n$. For $n=8$ tools and $\rho=0.5$, this means ~130 evaluations instead of 256.

\section{Experiments}

We evaluate AgentSHAP on the API-Bank benchmark \cite{li2023apibank}, which provides real executable tools and ground-truth annotations. We use GPT-4o-mini as the LLM and text-embedding-3-large for semantic similarity.

\subsection{Experimental Setup}

We use 8 tools from API-Bank: Calculator for arithmetic, QueryStock for stock prices, Wiki for Wikipedia search, plus AddAlarm, AddReminder, PlayMusic, BookHotel, and Translate. We use real queries from API-Bank's level-1 test set covering math calculations, stock price queries, and Wikipedia searches. We set sampling ratio $\rho = 0.5$ and run each experiment 3 times to measure consistency. We report Top-1 Accuracy (does the highest-SHAP tool match the expected tool), Cosine Similarity (how stable are SHAP vectors across runs), Quality Drop (response quality decrease when removing a tool), and SHAP Gap (difference between relevant and irrelevant tool scores).

\subsection{Consistency}

\begin{figure}[t]
    \centering
    \includegraphics[width=\columnwidth]{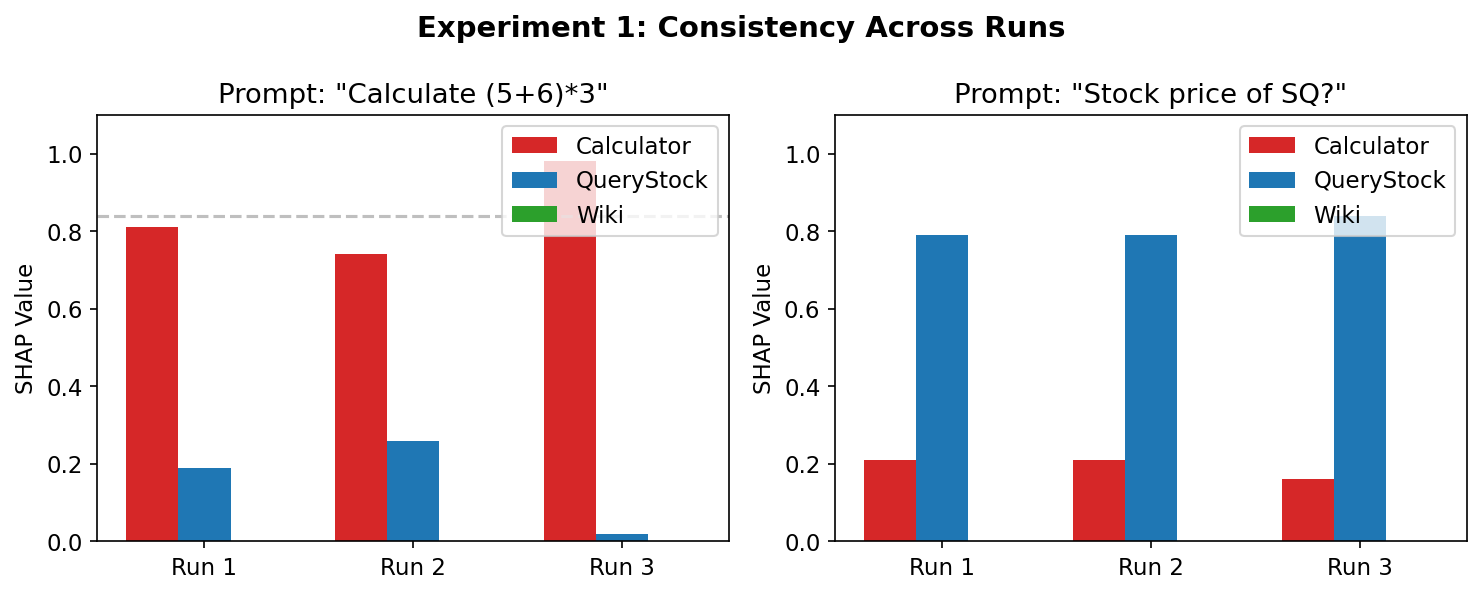}
    \caption{\textbf{Consistency across runs.} SHAP values are stable despite Monte Carlo sampling. Left: Math query consistently assigns high importance to Calculator. Right: Stock query consistently assigns high importance to QueryStock.}
    \label{fig:consistency}
\end{figure}

We test whether AgentSHAP results are stable across runs despite the stochastic nature of Monte Carlo sampling. We run AgentSHAP 3 times on 9 prompts with 3 tools available. Figure~\ref{fig:consistency} shows the results: for the math query "Calculate (5+6)*3", Calculator consistently receives the highest SHAP value (0.74-0.98) across all runs, and for stock queries, QueryStock consistently scores highest (0.79-0.84). The mean cosine similarity between SHAP vectors across runs is 0.945, indicating high stability, and top-1 accuracy is 100\%. This confirms Monte Carlo sampling provides stable estimates.

\subsection{Faithfulness}

\begin{figure}[t]
    \centering
    \includegraphics[width=\columnwidth]{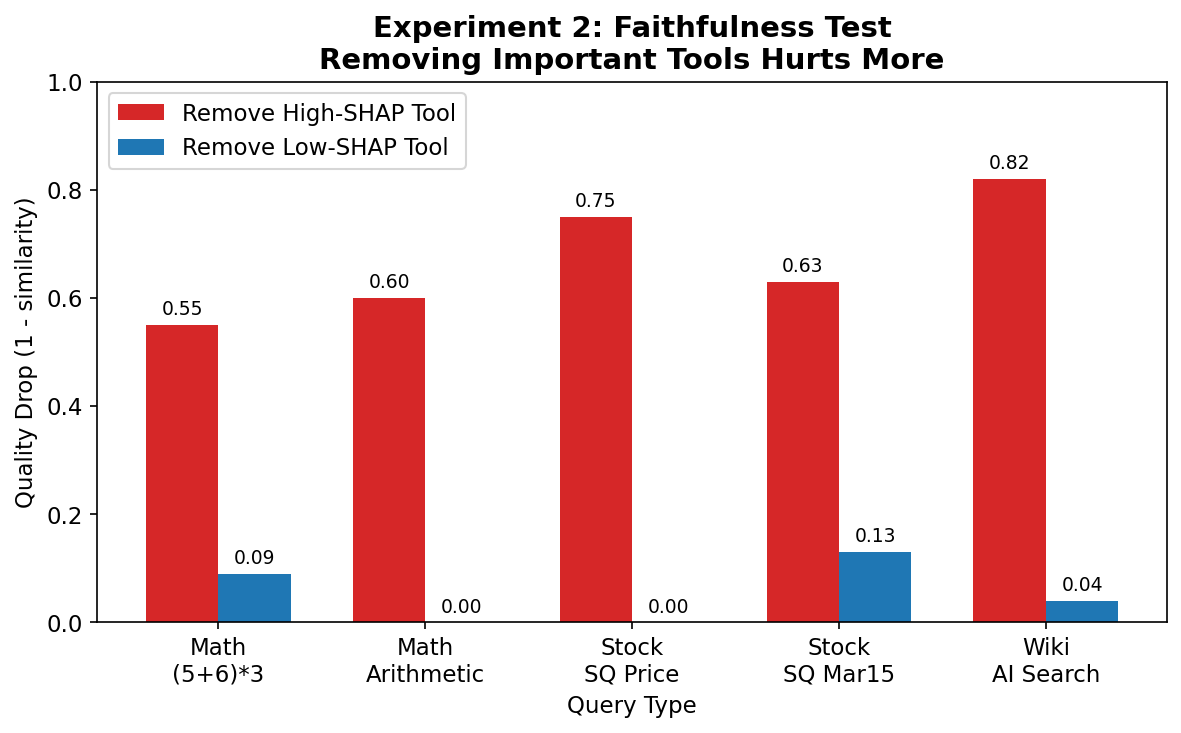}
    \caption{\textbf{Faithfulness test.} Removing high-SHAP tools (red) causes much larger quality drops than removing low-SHAP tools (blue). This confirms SHAP values reflect actual tool importance.}
    \label{fig:faithfulness}
\end{figure}

We test whether SHAP scores reflect actual tool importance by removing tools and measuring response quality change. For each prompt, we remove the highest-SHAP tool and the lowest-SHAP tool separately, then measure quality drop (semantic similarity to original response). Figure~\ref{fig:faithfulness} shows the results: removing high-SHAP tools causes mean quality drop of 0.67, while removing low-SHAP tools causes only 0.05 drop, a 13x difference. This confirms SHAP values accurately reflect which tools matter.

\subsection{Irrelevant Tool Injection}

\begin{figure}[t]
    \centering
    \includegraphics[width=\columnwidth]{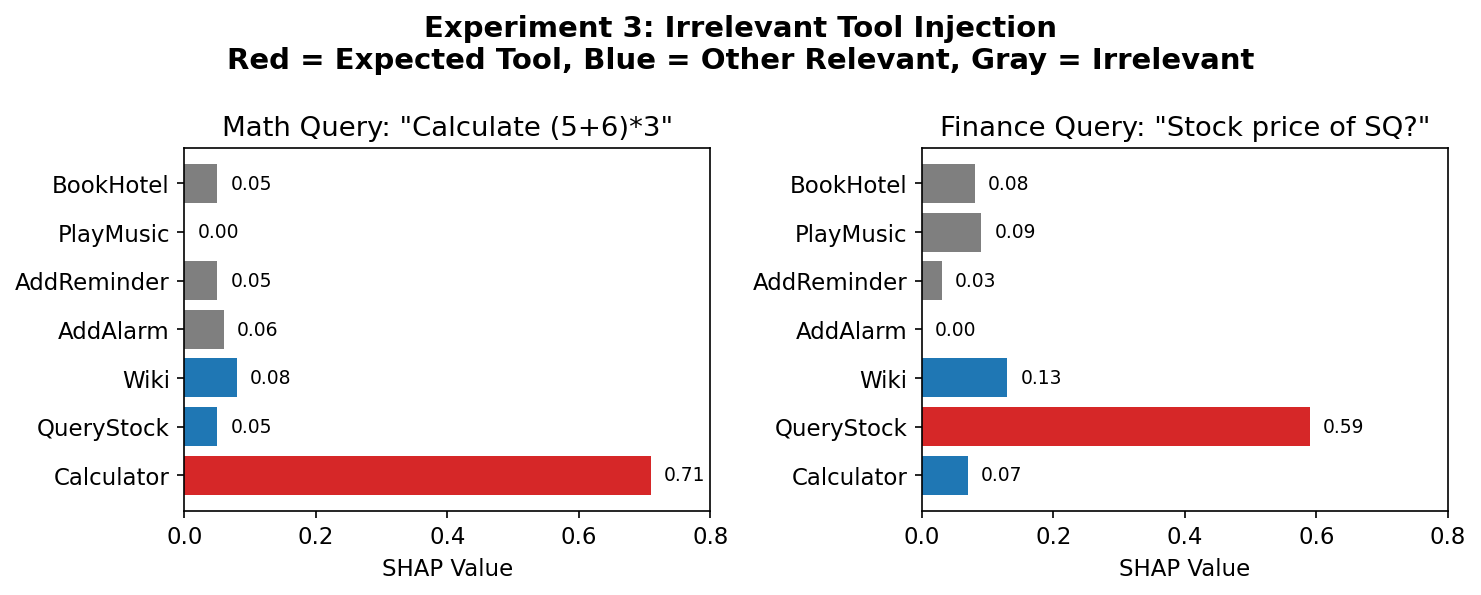}
    \caption{\textbf{Irrelevant tool injection.} AgentSHAP assigns low scores to injected irrelevant tools (gray). The expected tool (red) receives the highest score. Left: Math query. Right: Finance query.}
    \label{fig:injection}
\end{figure}

We test whether AgentSHAP can distinguish the actually-used tool from irrelevant tools by adding 4 unrelated tools (AddAlarm, AddReminder, PlayMusic, BookHotel) to the 3 core tools. Figure~\ref{fig:injection} shows the results: the expected tool (Calculator for math, QueryStock for finance, Wiki for knowledge) receives mean SHAP of 0.53 while irrelevant tools receive only 0.07, a 7x difference. Top-1 accuracy is 86\% (6/7 prompts correctly identify the expected tool). This demonstrates AgentSHAP can identify unnecessary tools for automatic pruning.

\subsection{Cross-Domain Attribution}

We test whether AgentSHAP correctly attributes importance across different query domains (Math, Finance, Knowledge) with 6 tools available. Figure~\ref{fig:crossdomain} shows the results as a heatmap: Math queries assign highest score to Calculator (0.45), Finance queries to QueryStock (0.55), and Knowledge queries to Wiki (0.57). Overall top-1 accuracy is 86\%, showing AgentSHAP correctly identifies domain-relevant tools.

\begin{figure}[t]
    \centering
    \includegraphics[width=\columnwidth]{images/fig_crossdomain.png}
    \caption{\textbf{Cross-domain attribution.} The heatmap shows mean SHAP values per domain-tool pair. Each domain correctly assigns highest importance to the expected tool: Calculator for Math (0.45), QueryStock for Finance (0.55), Wiki for Knowledge (0.57).}
    \label{fig:crossdomain}
\end{figure}

\subsection{Summary of Results}

Our experiments demonstrate that AgentSHAP provides reliable and meaningful tool importance scores. The consistency experiment shows that Monte Carlo sampling produces stable results with 0.945 mean cosine similarity between runs. The faithfulness experiment confirms that SHAP values reflect actual importance with 13x quality difference when removing important versus unimportant tools. The irrelevant tool injection experiment shows AgentSHAP assigns the expected tool 7x higher scores than irrelevant ones. The cross-domain experiment demonstrates 86\% accuracy across different query types. Together, these results validate AgentSHAP as a practical tool for understanding and optimizing LLM agent behavior.

\section{Limitations and Future Work}

AgentSHAP has several limitations that suggest directions for future research. First, it measures individual tool contributions without capturing synergies between tools. When Calculator and Wiki together produce better results than either alone, this interaction effect is not directly measured. Extending to pairwise or higher-order Shapley interactions could address this. Second, AgentSHAP analyzes single prompt-response pairs rather than multi-turn conversations where tool importance may shift over time. Tracking importance across conversation turns would provide richer insights for dialogue agents. Third, when agents call multiple tools in sequence, AgentSHAP attributes to the tool set rather than the specific calling order. Incorporating causal analysis of tool call chains could reveal ordering effects.

The main computational cost is LLM API calls, approximately $n + \rho \cdot 2^n$ calls for $n$ tools with sampling ratio $\rho$. For 8 tools with $\rho=0.5$, this means about 130 calls. Lower sampling ratios still provide useful estimates for cost-sensitive applications, but real-time explanations during agent execution remain challenging.

Future work could also use SHAP patterns to automatically recommend adding or removing tools, integrate with agent training pipelines for tool-aware fine-tuning, and extend to multi-agent systems where multiple agents collaborate using shared tools.

\section{Conclusion}

We presented AgentSHAP, the first framework for explaining tool importance in LLM agents. Using Monte Carlo Shapley values, AgentSHAP computes fair, theoretically grounded importance scores for each tool. Our experiments on API-Bank demonstrate strong results: 0.945 consistency across runs, 13x quality difference when removing important versus unimportant tools, 7x SHAP gap between the expected tool and irrelevant tools, and 86-100\% top-1 accuracy across experiments.

AgentSHAP enables practical applications including debugging agent errors, pruning unnecessary tools to reduce costs and latency, calibrating user trust, and A/B testing new tools. As LLM agents become ubiquitous, AgentSHAP provides the transparency needed to understand, debug, and trust their behavior.

\bibliography{references}

\end{document}